\documentclass[10pt, a4paper]{article}
\usepackage{lrec2026}
\usepackage{amsmath}
\usepackage{adjustbox}
\usepackage{multirow}
\usepackage{booktabs}   
\usepackage{caption}    
\usepackage{minted}
\newcommand{\baddelta}[1]{\textcolor{red!70!black}{+#1}}
\newcommand{\gooddelta}[1]{\textcolor{green!50!black}{\(-#1\)}}

\title{SloPal: A 60-Million-Word Slovak Parliamentary Corpus with Aligned Speech and Fine-Tuned ASR Models}

\name{Erik Božík\textsuperscript{1,4} \quad Marek Šuppa\textsuperscript{2,3,4}}

\address{
  \textsuperscript{1}VÚB Banka, Bratislava, Slovakia\\
  \textsuperscript{2}Faculty of Mathematics, Physics and Informatics, Comenius University in Bratislava\\
  \textsuperscript{3}Cisco Systems, Inc.~\textsuperscript{4}NaiveNeuron\\
  \textsuperscript{\textbf{Correspondence:}~\texttt{ebozik@vub.sk}}
}

\abstract{
Slovak remains a low-resource language for automatic speech recognition (ASR), with fewer than 100 hours of publicly available training data. We present SloPal, a comprehensive Slovak parliamentary corpus comprising 330,000 speaker-segmented transcripts (66 million words, 220 million tokens) spanning 2001--2024, with rich metadata including speaker names, roles, and session information. From this collection, we derive SloPalSpeech, a 2,806-hour aligned speech dataset with segments up to 30 seconds, constructed using a language-agnostic anchor-based alignment pipeline and optimized for Whisper-based ASR training. Fine-tuning Whisper on SloPalSpeech reduces Word Error Rate (WER) by up to 70\%, with the fine-tuned small model (244M parameters) approaching base large-v3 (1.5B parameters) performance at 6$\times$ fewer parameters. We publicly release the SloPal text corpus, SloPalSpeech aligned audio, and four fine-tuned Whisper models at \url{https://huggingface.co/collections/NaiveNeuron/slopal}, providing the most comprehensive open Slovak parliamentary language resource to date.
 \\ \newline \Keywords{Slovak speech recognition, Whisper fine-tuning, Parliamentary speech corpus, Low-resource ASR, Slavic languages} }

\begin{document}

\maketitleabstract
\section{Introduction} \label{sec:context}
Automatic Speech Recognition (ASR) has improved significantly with large multilingual datasets and models such as Whisper \cite{radford2023robust}. However, support for Slovak remains limited. To address this gap, we leverage publicly available parliamentary proceedings as a large-scale source of Slovak language data. We present SloPal, a 60-million-word corpus of speaker-segmented parliamentary transcripts, alongside SloPalSpeech, a derived aligned speech dataset, and fine-tuned ASR models.


Public ASR benchmarks, such as the Hugging Face Open ASR Leaderboard \cite{open-asr-leaderboard}, a snippet of which is shown in Table~\ref{tab:average_wer_top_models}, rank models by Word Error Rate (WER), the standard metric measuring the percentage of incorrectly recognized words. On English, where most evaluation is concentrated, top models achieve closely clustered results with minimal variation. In contrast, support for low-resource languages like Slovak is largely absent, leaving their performance unexplored. Moreover, many leading models either lack Slovak support entirely or are proprietary, restricting their usability for research.

\begin{table}
\centering
\small
\setlength{\tabcolsep}{4pt}
\begin{tabular}{l|r}
\toprule
\textbf{Model} & \textbf{Avg WER} \\
\midrule
nvidia/canary-qwen-2.5b & 5.63 \\
ibm-granite/granite-speech-3.3-8b & 5.85 \\
nvidia/parakeet-tdt-0.6b-v2 & 6.05 \\
microsoft/Phi-4-multimodal-instruct & 6.14 \\
nvidia/canary-1b-flash & 6.35 \\
kyutai/stt-2.6b-en & 6.40 \\
nvidia/canary-1b & 6.50 \\
nyrahealth/CrisperWhisper & 6.67 \\
ibm-granite/granite-speech-3.3-2b & 6.86 \\
elevenlabs/scribe\_v1 & 6.88 \\
speechmatics/enhanced & 6.91 \\
\bottomrule
\end{tabular}
\vspace{2mm}
\caption{Average WER (\%) on English test sets of top-rated multilingual models in Open ASR leaderboard at the time of writing}
\label{tab:average_wer_top_models}
\end{table}

Models like OpenAI's Whisper \cite{radford2023robust} rely on large, diverse datasets, yet public resources remain scarce for many languages, leading to weaker performance. As shown in Table~\ref{tab:wer_low_resource}, Whisper achieves much lower WERs on high-resource languages (e.g., English, Russian, German; thousands of training hours) than on low-resource ones (e.g., Slovak, Croatian, Lithuanian; $\sim$100 training hours). Large models achieve 3--5\% WER on high-resource languages but remain above 10\% on low-resource ones---a gap driven primarily by data scarcity rather than model capacity.

Recent work has demonstrated that fine-tuning Whisper on domain-specific or language-specific data can substantially improve performance. \citet{liu2024exploration} conducted systematic experiments comparing various parameter-efficient fine-tuning approaches for Whisper, showing that targeted adaptation can yield significant improvements even with limited computational resources. \citet{timmel2024fine} explored fine-tuning strategies for low-resource settings, confirming that even modest amounts of in-domain data can substantially reduce WER when properly leveraged.

\subsection{Parliamentary Speech Corpora for ASR}
Parliamentary proceedings have emerged as a valuable source of speech data for ASR across multiple languages, offering long-form recordings with corresponding transcripts, formal register with clear articulation, and substantial volume. \citet{helgadottir2017building} pioneered this approach with the Icelandic Althingi corpus, collecting 542 hours of parliamentary speech and establishing methodological foundations for alignment and segmentation of long-form recordings. \citet{solberg2022norwegian} developed the Norwegian Parliamentary Speech Corpus (NPSC), comprising 140 hours of speech, demonstrating a 22.9\% relative WER reduction. \citet{kulebi2022parlamentparla} introduced ParlamentParla, a 600-hour corpus of Catalan parliamentary speech, providing insights into handling speaker diarization and long-form alignment challenges. \citet{rekathati2023rixvox} released RixVox, a Swedish parliamentary corpus with over 5,400 hours of speech, representing the largest single-language parliamentary dataset to date. \citet{virkkunen2023finnish} released the Finnish Parliament ASR corpus with over 3,000 hours of speech. \citet{stankov2025parczech} created ParCzech4Speech with 2,695 hours of Czech speech, a West Slavic language closely related to Slovak. More recently, \citet{ljubesic2024parlaspeech} released the ParlaSpeech collection, an ambitious multi-lingual effort encompassing Croatian (3,110 hours), Polish (1,010 hours), and Serbian (896 hours), demonstrating the scalability of parliamentary corpus construction across Slavic languages. On the textual side, the ParlaMint project \cite{erjavec2023parlamint} created uniformly encoded, linguistically annotated parliamentary corpora for 29 European parliaments, though notably without Slovak or aligned audio. Table~\ref{tab:parliamentary_comparison} summarizes these parliamentary speech corpora alongside SloPalSpeech.

\begin{table}[!t]
\centering
\resizebox{\columnwidth}{!}{%
\begin{tabular}{l|l|r}
\toprule
\textbf{Corpus} & \textbf{Language} & \textbf{Hours} \\
\midrule
Althingi \cite{helgadottir2017building} & Icelandic & 542 \\
NPSC \cite{solberg2022norwegian} & Norwegian & 140 \\
ParlamentParla \cite{kulebi2022parlamentparla} & Catalan & 600 \\
RixVox \cite{rekathati2023rixvox} & Swedish & 5,493 \\
Finnish Parl. \cite{virkkunen2023finnish} & Finnish & 3,130 \\
ParCzech4Speech \cite{stankov2025parczech} & Czech & 2,695 \\
ParlaSpeech-HR \cite{ljubesic2024parlaspeech} & Croatian & 3,110 \\
ParlaSpeech-PL \cite{ljubesic2024parlaspeech} & Polish & 1,010 \\
ParlaSpeech-SR \cite{ljubesic2024parlaspeech} & Serbian & 896 \\
\midrule
EuroSpeech \cite{pfisterer2025eurospeechmultilingualspeechcorpus} & 22 languages & 50,500 \\
\quad \textit{EuroSpeech-Slovak subset} & \textit{Slovak} & \textit{2,554} \\
\midrule
\textbf{SloPalSpeech (ours)} & \textbf{Slovak} & \textbf{2,806} \\
\bottomrule
\end{tabular}%
}
\vspace{2mm}
\caption{Comparison of parliamentary speech corpora for ASR. EuroSpeech is a multilingual corpus with 22 languages totaling 50,500 hours (CER $<$ 20\%), including 2,554 hours for Slovak.}
\label{tab:parliamentary_comparison}
\end{table}

At 2,806 hours, SloPalSpeech is the first dedicated Slovak parliamentary speech resource, comparable in scale to ParCzech4Speech (2,695 hours for Czech, a closely related West Slavic language) and the Slovak subset of the concurrent EuroSpeech corpus (2,554 hours). The broader SloPal corpus additionally provides the complete textual record---330,000 speaker-segmented transcripts---making it the largest Slovak parliamentary language resource in any modality.

\subsection{Relation to EuroSpeech}
Concurrent with our work, \citet{pfisterer2025eurospeechmultilingualspeechcorpus} introduced EuroSpeech, a multilingual parliamentary corpus including 2,554 hours of Slovak speech using a two-stage dynamic alignment achieving $\sim$89\% coverage. While EuroSpeech excels at audio alignment across 22 languages, SloPal provides complementary strengths: (1) comprehensive textual coverage\footnote{By textual coverage we mean that SloPal includes the full text of all successfully retrieved and parsed transcripts, independent of whether the corresponding audio could be aligned. Some transcript files were unavailable due to download errors (see Section~3.3), and audio recordings cover only forenoon sessions (see Section~3.5).} (66M words in 330,000 speaker-segmented transcripts vs.\ 11,700 segments in EuroSpeech's publicly available Slovak subset), (2) rich speaker metadata enabling discourse analysis, (3) 30-second segment lengths optimized for Whisper's architecture, and (4) Slovak-specific fine-tuned models. Our dual-resource design provides complete text coverage for NLP applications while offering 2,806 hours of aligned audio for ASR training. Both resources draw from the same Slovak parliamentary sources (MediaPortál), serving distinct but complementary research needs: EuroSpeech for multilingual ASR benchmarking, SloPal for comprehensive Slovak language and discourse research. We note that Slovak parliamentary audio is publicly available and portions may also appear in other large-scale datasets such as VoxPopuli \cite{wang2021voxpopuli}, which includes European Parliament recordings; SloPal is the first resource to systematically cover the full Slovak National Council proceedings with speaker-level segmentation.

\vspace{-2mm}
\begin{table*}[!t]
\centering
\setlength{\tabcolsep}{3pt}
\footnotesize
\begin{tabular}{l|rrr|rrr}
\toprule
\textbf{Model}
& \multicolumn{3}{c|}{\textbf{Low-resource}} & \multicolumn{3}{c}{\textbf{High-resource}} \\
& Slovak & Croatian & Lithuanian & English & Russian & German \\
\midrule
Whisper tiny     & 77.2 & 79.0 & 98.5 & 12.4 & 31.1 & 27.8 \\
Whisper small    & 33.3 & 33.4 & 65.6 &  6.1 & 11.4 & 10.2 \\
Whisper medium   & 17.3 & 19.3 & 41.1 &  4.4 &  7.2 &  6.5 \\
Whisper large‑v2 & 11.7 & 13.4 & 28.1 &  4.2 &  5.6 &  4.5 \\
\midrule
\textbf{Hours} & \textbf{90} & \textbf{91} & \textbf{67} & \textbf{438k} & \textbf{9.8k} & \textbf{13.3k} \\
\bottomrule
\end{tabular}
\caption{Whisper WERs (\%) on FLEURS for low-resource vs.\ high-resource languages, illustrating the gap driven by training data availability}
\label{tab:wer_low_resource}
\vspace{-3mm}
\end{table*}

\section{Objectives}
The objective of this work is to address the challenges faced by low-resource languages, with particular focus on Slovak. Our primary goal is to create comprehensive resources from parliamentary proceedings: (1) SloPal, a complete corpus of speaker-segmented transcripts capturing the full textual record, and (2) SloPalSpeech, an aligned audio-transcript dataset suitable for ASR model training with segments up to 30 seconds.

Since publicly available resources rarely exist in aligned format, preprocessing inevitably loses some data during alignment. We address this by releasing both the complete textual corpus (SloPal, 60 million words) and the successfully aligned subset (SloPalSpeech, 2,806 hours), maximizing resource utility.

The 30-second segment length aligns with the Whisper family of models, which we fine-tune using SloPalSpeech. Our goal is to develop Slovak-optimized ASR models spanning sizes from Whisper-small to Whisper-large-v3, assessing the impact of parliamentary data on model performance and establishing benchmarks for Slovak ASR.

\section{Dataset}

Several open-source datasets exist for Slovak, such as Common Voice \cite{ardila2019common}, a crowd-sourced multilingual speech corpus, FLEURS \cite{conneau2023fleurs}, a multilingual speech benchmark derived from translated Wikipedia sentences, and VoxPopuli \cite{wang2021voxpopuli}, a large-scale multilingual speech dataset based on European Parliament recordings. However, together they offer only about 100 hours of Slovak data, which is far below the scale typically required for high-performance ASR.

We focus on Slovak parliamentary hearings to bridge this gap, a source that yields thousands of hours of training data, surpassing existing Slovak datasets by an order of magnitude. These hearings are fully public and regularly transcribed, making them a strong candidate for ASR dataset creation. Audio recordings and their corresponding transcripts are publicly available on various government websites, with coverage dating back to the establishment of the Slovak Republic. This long-term availability presents a valuable opportunity to construct a dataset that qualifies as large-scale.

\subsection{MediaPortál} \label{sec:mediaportal}

Parliamentary recordings and transcripts are available across multiple platforms. While older sessions are accessible through a legacy portal with inconsistent coverage, MediaPortál \cite{mediaportal_nr_sr} has emerged as the primary and most comprehensive source for more recent election terms. It provides a centralized interface for browsing and accessing audio recordings of parliamentary hearings, making it the most suitable foundation for building our ASR dataset.

\begin{figure}
    \centering
    \includegraphics[width=1\linewidth]{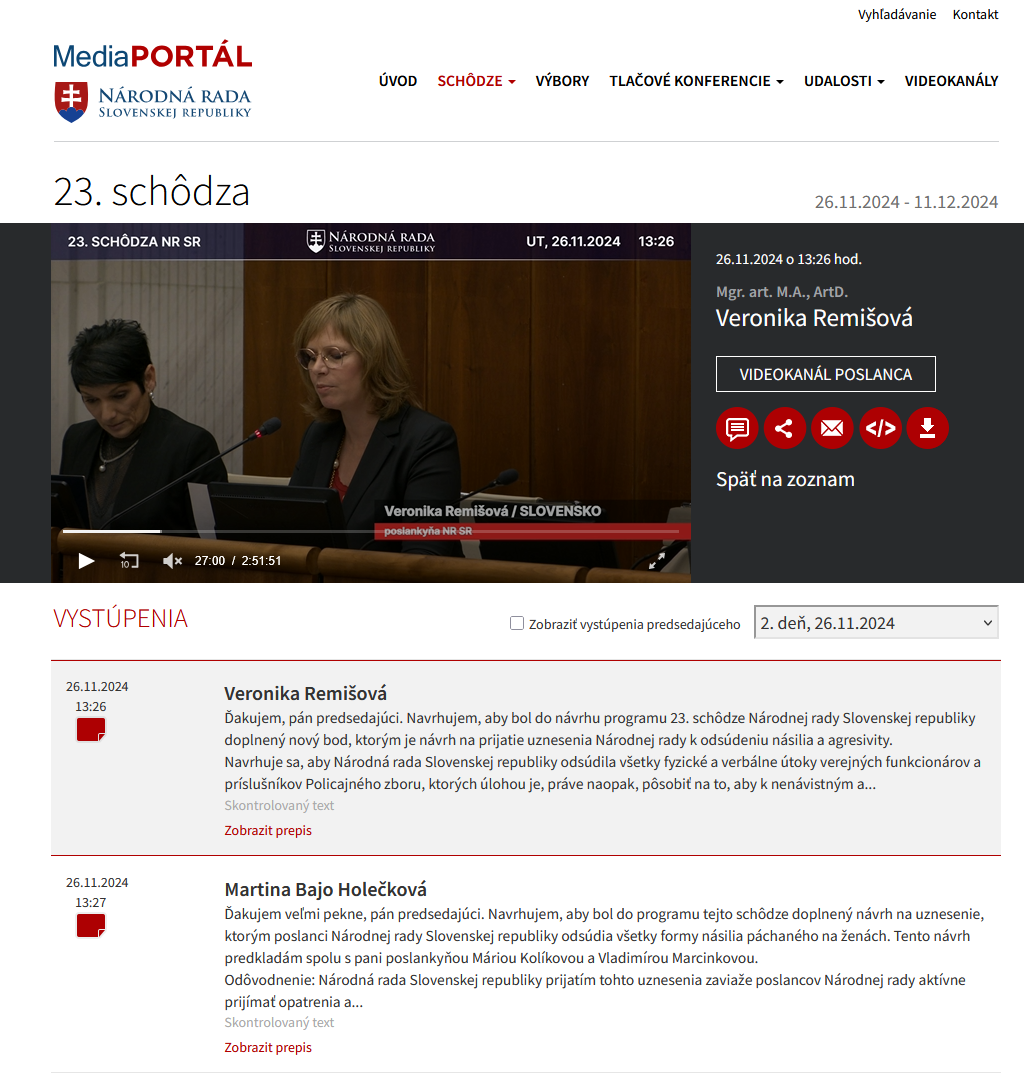}
    \caption{Sample page from MediaPortál}
    \label{fig:mediaportal}
\end{figure}

Hearings in MediaPortál follow a structure similar to the one shown in Figure~\ref{fig:mediaportal}. From the layout, users can navigate through the entire recording by simply clicking on the speaker segments. However, we observed that these segments were often inaccurate. Due to these inconsistencies, we concluded that, in order to construct a dataset based on our predefined structure, the data would need to be re-aligned. As a result, we abandoned the idea of directly using the provided segments and instead focused on collecting the full-text transcript. While it is possible to generate the transcript by concatenating all individual segments, we opted to use the official administrative version of the transcript, which is readily available in full-text form on The Joint Czech and Slovak Digital Parliamentary Library \cite{nrsr_dl}.

\subsection{The Joint Czech and Slovak Digital Parliamentary Library}
This initiative, established in 2002 in Slovakia, aims to digitize parliamentary documents. Since verbatim transcripts of parliamentary hearings fall within this category, these documents can be easily filtered and accessed in digital form through the portal. The platform features a tree-view interface, which offers an intuitive way to filter transcripts for each session.
Under each session, multiple links are provided - each corresponding to the verbatim transcript of a specific meeting day, available in DOCX format.

\begin{figure}
    \centering
    \includegraphics[width=1\linewidth]{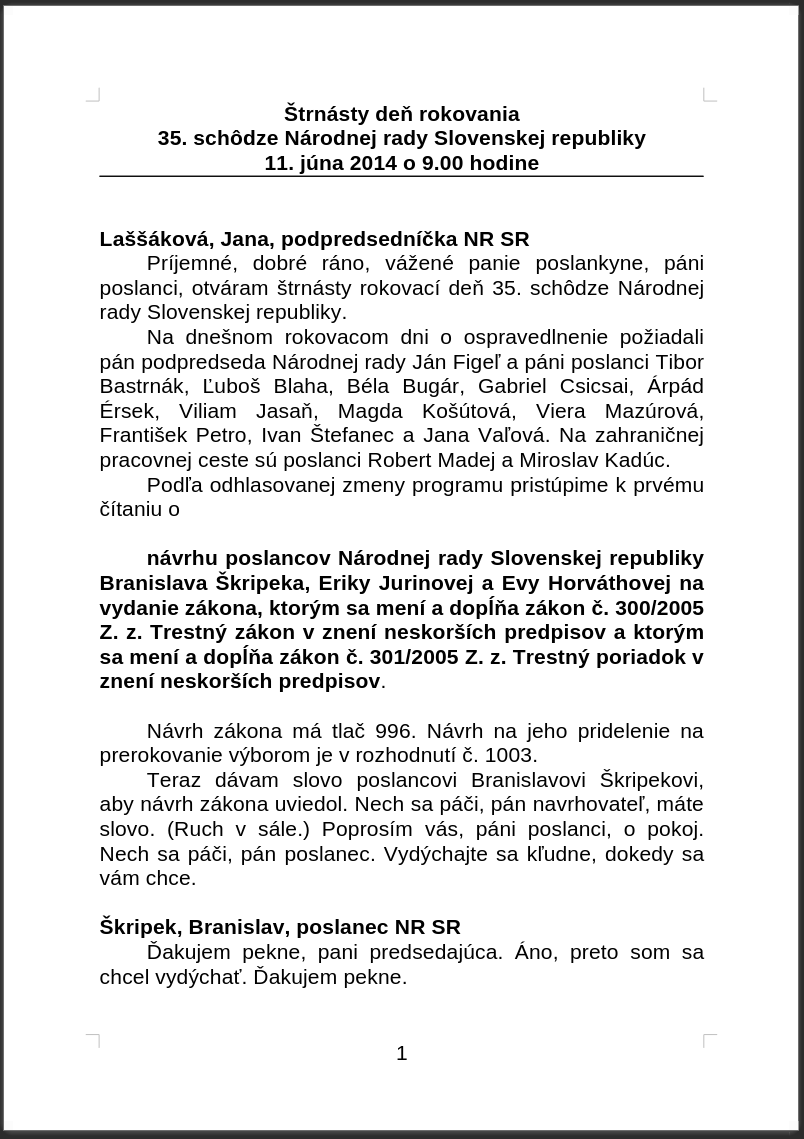}
    \caption{Example transcript in .docx format}
    \label{fig:example_transcript}
\end{figure}

Our examination of transcripts across different years revealed that documents which cover all the session recordings available in MediaPortál follow a consistent formatting structure, as shown in Figure~\ref{fig:example_transcript}. The layout is organized around speaker announcements, which are always presented in bold. These announcements follow a standardized format: surname, first name, and the speaker’s role at the time, each separated by a comma. Bold text is also used beyond speaker announcements - for instance, in headers that convey important information such as the date and time of the session. Between each speaker announcement appears the actual verbatim transcript of their speech.


\subsection{Collection}


We collected all available audio recordings from MediaPortál. Since the frontend download functionality was not working, we extracted the URLs of the HLS (HTTP Live Streaming) streams and used FFmpeg \cite{ffmpeg_doc} to download the recordings. Due to storage limitations, we opted to extract only the audio in MP3 format. This resulted in a total of 4096 hours of audio.

For the transcripts, we directly downloaded the available DOCX documents. Some documents had to be discarded because the webpage redirected to an HTML page instead of a valid DOCX file. In total, we successfully collected 2199 transcript files.

To match each audio recording with the corresponding transcript, we extracted both the session number and the exact date. This was necessary because a single session number can span multiple days. As a result, the session number alone was not sufficient to produce an accurate match. Therefore, the date was also required to uniquely identify each recording. Using this method, we were able to match 1136 recordings, covering a total of 4098 hours of audio.

\subsection{Parsing}

The first step in effectively parsing DOCX files was to convert them into a format that is easier to process. For this purpose, we used Apache Tika (specifically the Tika Server) \cite{apache_tika_wiki} which enabled conversion to XHTML. This transformation allowed us to perform HTML-like parsing, as long as XHTML-specific features were disregarded. As a result of the conversion, some formatting elements were preserved as tags, most notably bold text enclosed in \texttt{<b></b>} tags.

To filter out all non-verbatim content, we needed to segment the transcripts based on speaker annotations. However, due to inconsistencies introduced during manual transcription (e.g., missing commas, inconsistent formatting of names) and edge cases such as multi-part names, simple rule-based splitting using punctuation or regular expressions proved unreliable. We therefore adopted a heuristic-based approach for identifying speaker annotations using the following conditions:

\begin{enumerate}
    \item The text was \textbf{bold}.
    \item It contained \textbf{at least one and no more than three names}.
    \item It \textbf{did not exceed 15 words}.
\end{enumerate}

These thresholds reflect the structure of Slovak parliamentary transcripts: speaker annotations typically contain a surname, first name, and role designation (hence one to three name tokens), and never exceed a short line (hence the 15-word limit), while longer bold passages invariably correspond to section headers or emphasized speech.

To support this logic, we scraped a complete list of National Council members from all electoral terms using the official parliamentary website \cite{nrsr_poslanci_abc}. From this data, we compiled a comprehensive \textbf{set of known names}, including both first and last names for each member. Each time a bold line was encountered, we checked it against this set. If the required number of name matches was found, the line was classified as a speaker annotation. During validation, we observed a small number of mismatches caused by missing names in the scraped list, primarily names of Slovak presidents and government ministers who address parliament but are not members. These were added manually to ensure correct speaker identification. Other occasional non-MP speakers (e.g., invited experts) appear rarely in the transcripts and are handled by the same heuristics, since their annotations follow the same bold-text format with name and role.

These heuristics were developed iteratively through manual inspection of the parsing outputs. During development, we repeatedly sampled parsed transcripts and adjusted the conditions until the resulting segmentation produced consistent speaker annotations without introducing obvious errors. In particular, we introduced the word limit after observing that some bold segments matched the name-based conditions but corresponded to emphasized spoken text rather than speaker identifiers.

Additionally, we filtered out all text enclosed in parentheses, as such segments typically represent transcriber's notes (e.g., descriptions of non-verbal events). Specifically, we removed all text enclosed in both \texttt{()} and \texttt{[]}. To handle cases where parentheses were not properly closed, we implemented a fallback rule that treats a bracket as the start of a note and the next period as its end. After applying these steps, the transcripts were transformed into a semi-structured format consisting of alternating speaker annotations and corresponding speech segments.

\subsection{Validity}
To ensure the transcripts matched the audio, we randomly sampled five sessions and created a test set. We then evaluated several Whisper models on the audio and measured the WER against the provided normalized transcripts.\footnote{We used the WhisperX framework \cite{bain2023whisperx} for its ability to process long-form audio, acknowledging that results may differ slightly with alternative splitting methods or without the CTranslate2 Whisper implementation. Our goal was to validate the transcripts rather than precisely evaluate model performance.} Initial WER values exceeded 1.0, indicating a major mismatch. This led us to discover that the audio only included the forenoon part of sessions, while the transcripts covered both forenoon and afternoon. After removing the afternoon parts from the transcripts, we obtained WERs shown in Table~\ref{tab:wer-whisper}, which closely match those reported by the Whisper authors. We therefore consider the transcripts valid for the forenoon parts.

\begin{table}[ht]
\centering
\small
\setlength{\tabcolsep}{10pt}
\begin{tabular}{l|r}
\toprule
\textbf{Model} & \textbf{Word Error Rate} \\
\midrule
openai/whisper-\textbf{large-v3} & 12.9\% \\
openai/whisper-\textbf{medium}   & 21.7\% \\
openai/whisper-\textbf{small}    & 36.1\% \\
openai/whisper-\textbf{tiny}     & 77.3\% \\
\bottomrule
\end{tabular}
\vspace{2mm}
\caption{WER of different Whisper models (their CTranslate2 implementation) on parliamentary transcripts}
\label{tab:wer-whisper}
\end{table}

\section{Alignment and segmentation}

The collected data consisted of transcripts and audio recordings of varying lengths, ranging from a few seconds up to 25 hours, with an average duration of 3.6 hours. This diverged significantly from our target structure, which limits segments to a maximum of 30 seconds. As noted in Section~\ref{sec:mediaportal}, this made both alignment and segmentation necessary to meet our required dataset structure.

Although forced alignment is a common method for generating timestamps, it was not suitable in our case. The transcripts frequently included more content than was actually spoken, and the long duration of the recordings exceeded the effective input range of most forced alignment models, which typically perform better on shorter segments. To overcome these limitations, we developed an alternative approach based on generating a reference transcript from the audio and aligning it with the collected transcript, which we treated as the ground truth.

\subsection{Reference Transcript}

We first generated a reference transcript with word-level timestamps using the WhisperX framework, which combines Whisper’s transcription capabilities with external alignment models \cite{bain2023whisperx}. While Whisper provides accurate results (with a WER of 12.9\% for the largest model, as shown in Table~\ref{tab:wer-whisper}), the output still contained typical inaccuracies. However, since it outputs transcripts in 30-second chunks, it enabled effective forced alignment afterward without the limitations previously mentioned.

\subsection{Anchor-Based Timestamping}

To align the reference transcript with the ground truth, we identified “anchors” - words that appear in both versions and can serve as reliable reference points. We assumed that some words in the reference were correctly transcribed and could be matched to their counterparts in the ground truth. Each reference word was matched to candidate ground truth words within a Levenshtein distance of 1 \cite{levenshtein1966binary}, provided the match did not jump backward relative to the last aligned word and stayed within a 50-word forward window. If no valid candidates were found, the word was skipped.

Among all candidates, we selected the best match by comparing the context: specifically, the four preceding and four following words. Each candidate received a score based on the number of full-text matches in this window, and the one with the highest score was chosen as the anchor. To further improve accuracy, we applied several heuristics, such as skipping words shorter than three characters and requiring a minimum match score of 3 for a word to qualify as an anchor.

This process yielded a list of ground truth words with corresponding timestamps. Because each match was constrained to move forward, the resulting timestamps formed a strictly increasing sequence, which effectively eliminated alignment anomalies.

Importantly, this anchor-based approach is language-agnostic: it requires only a Whisper model capable of producing approximate transcripts in the target language and a ground-truth transcript. Because Whisper supports over 90 languages, the method can be applied to parliamentary corpora in any of those languages without training a dedicated forced-alignment model.

\subsection{Segment Construction from Anchors}

The anchoring method was effective: most sessions contained a sufficient number of matched words. To construct segments suitable for our ASR dataset, we aimed to group transcript and audio pairs that were no longer than 30 seconds.

We iterated through the list of anchors, starting from each one and scanning forward until we found the next anchor with a time difference greater than 28 seconds. We used a 2-second buffer to create a soft threshold. All ground truth words between the starting anchor and this later anchor formed a candidate segment, and the corresponding audio slice was extracted using their timestamps.

This approach yielded a useful number of segments across most sessions. However, in cases with sparse anchors - caused by long pauses or local misalignments - a segment was still constructed, but its validity required further assessment.

\subsection{Final Selection}

\begin{figure}[ht]
    \centering
    \includegraphics[width=\linewidth]{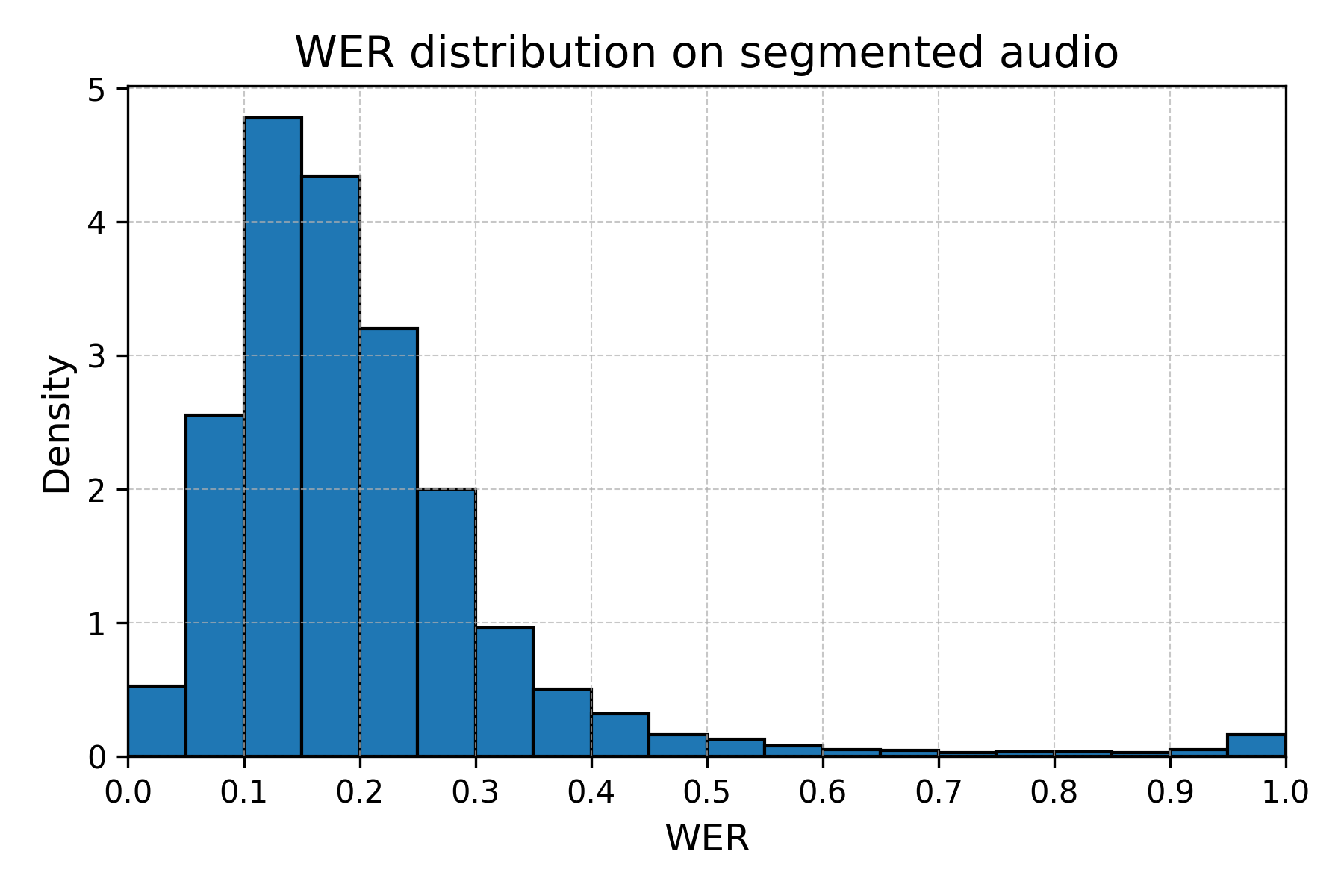}
    \caption{Distribution of WER values on segmented audio}
    \label{fig:wer_segmented}
\end{figure}

Although the algorithm itself was deterministic, it operated on Whisper-generated transcripts, which are inherently variable and sometimes unpredictable. This occasionally led to anomalies or undesired behavior in specific contexts. To address this, we applied a post-processing step where the Whisper model was reapplied to each segmented audio chunk shorter than 30 seconds\footnote{This time without WhisperX since Whisper could be applied directly - enabling faster inference}, generating updated transcripts and subsequently calculating the WER. The resulting distribution is shown in Figure~\ref{fig:wer_segmented}. While most segments had a WER below 50\%, a notable portion exhibited significantly higher values. We analyzed different WER intervals and found that lower rates typically indicated well-aligned and correctly segmented audio, whereas higher WERs were often associated with missegmentations or transcription errors. This trend was confirmed through manual inspection across the range of values.

To remove problematic segments while preserving as much usable content as possible, we empirically selected a WER threshold of 40\%. Applying this filter reduced the dataset from 2977.73 to 2806 hours, eliminating 171.73 hours of low-quality data. In terms of segments, the count decreased from 427{,}276 to 402{,}966 - a reduction of 24{,}310 segments. This threshold effectively retained the majority of the data while substantially improving its overall quality.

\section{Fine-tuning}
With the dataset prepared, we fine-tuned multiple Whisper models to measure the impact of parliamentary speech data on Slovak ASR performance. This section outlines our experimental setup and the fine-tuning strategies adopted for different model sizes, including the adaptations required for the largest variant.

\subsection{General Setup}

We utilized multiple NVIDIA A10 GPUs to accelerate training. Given that Whisper has been widely adopted since its 2022/2023 release, we implemented the fine-tuning pipeline using Hugging Face Transformers, which simplifies much of the low-level implementation. Our fine-tuning strategy involved training for a maximum of three epochs with early stopping, based on evaluation performance. This setup was informed by prior research on transformer models such as \cite{Devlin2018}, as well as the original Whisper pre-training protocol \cite{radford2023robust}, which also relied on a small number of epochs.

Although we had over 2,800 hours of training data, its formal and domain-specific nature raised concerns about catastrophic forgetting. To assess generalization, we monitored performance during training on held-out evaluation sets: a parliamentary set (~300 samples) and 300-sample subsets from FLEURS and Common Voice. We selected subsets of 300 samples to keep evaluation computationally feasible without significantly slowing down training. These subsets were sampled from the official test splits; however, they were used solely for early stopping decisions at epoch-level granularity and never for gradient updates. Final benchmarks in Section~6 were conducted on the complete test sets.

\subsection{Whisper Small, Medium, Turbo}

Fine-tuning these models was feasible on a single GPU due to their manageable size. We based our code on a blog post by \cite{gandhi2022fine} and adopted most of the original hyperparameters, with the exception of the learning rate. While many works default to a fixed value of $1 \times 10^{-5}$, we followed a recommendation by \cite{lodagala2023whisperfinetune} of using a learning rate 40× smaller than that used during pretraining. In our experiments, this adjustment led to more stable convergence across most of the model sizes.

We applied early stopping between the 2nd and 3rd epoch after observing evaluation plateaus. None of the models showed signs of overfitting or catastrophic forgetting. All achieved substantial improvements on the parliamentary evaluation set and importantly also generalized well to broader Slovak speech, yielding notable WER reductions on both FLEURS and Common Voice during training.

\subsection{Whisper Large v3}
Fine-tuning Whisper Large v3 required adjustments beyond those used for smaller models, mainly due to its size exceeding single-GPU memory limits. While official details on its training are unavailable, it is known to have been trained on millions of hours of data and to use a differently sized Mel spectrogram than earlier versions. These differences initially led to configuration issues and overfitting, which we had to address.

\subsubsection{Multi-GPU setup}
 We addressed the GPU limitations by using Fully Sharded Data Parallel (FSDP) \cite{zhao2023pytorch}. Although Hugging Face Transformers \cite{wolf-etal-2020-transformers} officially support FSDP and multi-GPU training via Accelerate \cite{accelerate}, we faced persistent compatibility issues - different library versions produced inconsistent runtime errors with the same code. Training was eventually stabilized, but evaluation consistently failed due to runtime errors.\footnote{Stable training required Transformers v4.52.4; evaluation required a separate process due to FSDP-related runtime errors.} As a workaround, we used three GPUs for training and a fourth GPU to run inference at selected checkpoints. This differed from the single-GPU setup used for smaller models, where training and evaluation could run together in a single process.

We used a sharded state dict configuration, which produced FSDP checkpoints during training. These were merged into a single \texttt{safetensors} file, which we then converted back to its original format with split weights, matching the structure used in the original release.

\subsubsection{Overfitting}
Starting with the same setup as previous models, we used a learning rate 40× smaller, as mentioned earlier. However, after the warmup stage, WER quickly increased - first on the parliamentary test set, then on the other evaluation sets. This indicated significant overfitting, which needed to be addressed to obtain a usable model. As noted in forum discussions \cite{whisper_finetuning_2024}, others reported encountering similar issues, though no definitive solution was identified. We reverted to the commonly reported settings, excluding LoRA adapters, as LoRA was not our objective. This included adjusting the learning rate to $1 \times 10^{-5}$. After this we proceeded with hyperparameter experimentation. Introducing a weight decay of 0.01 (previously 0) enabled stable convergence in stages that had previously overfitted, reducing WER across all test sets. This became our final configuration for Whisper Large v3 with which we proceeded until an early stopping around 2 epochs.

\section{Benchmarks}
To evaluate the impact of parliamentary speech fine-tuning on Slovak ASR performance, we conducted a comprehensive benchmarking of all produced models. While preliminary indicators were available from the training process, final assessments were carried out on the official test sets of Common Voice 21 and FLEURS. As with all benchmarking, we included basic text normalization before producing the WER value.

\subsection{Results}



\begin{table}[ht]
\centering
\small
\setlength{\tabcolsep}{4pt}
\begin{tabular}{l|cl|cl}
\toprule
\textbf{Model} & \multicolumn{2}{c|}{\textbf{CV21 (\%)}} & \multicolumn{2}{c}{\textbf{FLEURS (\%)}} \\
\midrule
Small-Base      & 58.4 & & 36.1 & \\
Small-Tuned     & 25.7 & (\gooddelta{32.7}) & 10.6 & (\gooddelta{25.5}) \\
\midrule
Medium-Base     & 38.0 & & 18.7 & \\
Medium-Tuned    & 18.0 & (\gooddelta{20.0}) & 7.6 & (\gooddelta{11.1}) \\
\midrule
Turbo-v3 Base   & 31.7 & & 10.7 & \\
Turbo-v3 Tuned  & 13.2 & (\gooddelta{18.5}) & 6.4 & (\gooddelta{4.3}) \\
\midrule
Large-v3 Base   & 20.8 & & 9.2 & \\
Large-v3 Tuned  & 11.6 & (\gooddelta{9.2}) & 5.5 & (\gooddelta{3.7}) \\
\bottomrule
\end{tabular}
\vspace{2mm}
\caption{WER comparison of Whisper models (Base vs fine-tuned). Improvements are shown in parentheses as absolute WER reduction in percentage points.}
\label{tab:wer-results}
\end{table}

The benchmarking outcomes on the whole test sets of Common Voice 21 and FLEURS, presented in Table~\ref{tab:wer-results}, show WER decreases across all models after fine-tuning, with the largest relative improvements observed for the smaller models.
Fine-tuning Whisper Small reduced WER by 65--70\%, bringing a 244M-parameter model within striking distance of the 1.5B-parameter base Large-v3. Gains diminish with model size, indicating that approximately 2,800 hours of in-domain data most benefits smaller, deployment-friendly models. Manual inspection confirmed that fine-tuning improved both word-level accuracy and the generation of Slovak-like token compositions, even when predictions were partially incorrect. Whisper Medium showed similar but smaller improvements, still yielding substantial performance gains.

Despite Whisper Large-v3 being trained on millions of hours of data, our dataset further reduced WER by a few percentage points. At these low error rates, improvements were less apparent during manual inspection, though handling of rare Slovak words noticeably improved. Among all evaluated models, Whisper Large-v3-Turbo offers the best efficiency trade-off: it matches Large-v3 accuracy within 1 percentage point while using 730M fewer parameters, making it the recommended choice for production Slovak ASR deployment.

Table~\ref{tab:wer-large-canary-parakeet} compares our largest fine-tuned model with NVIDIA's Canary-1B-v2 and Parakeet-TDT-0.6B-v3 \cite{sekoyan2025canary1bv2parakeettdt06bv3efficient}. We observe improvements over Canary on both benchmarks. Note that FLEURS and Common Voice v7 were included in the NVIDIA models' training data, giving them an advantage on these test sets.

\begin{table}[t]
\centering
\small
\setlength{\tabcolsep}{4pt}
\begin{tabular}{l|cl|cl}
\toprule
\textbf{Model} & \multicolumn{2}{c|}{\textbf{CV21 (\%)}} & \multicolumn{2}{c}{\textbf{FLEURS (\%)}} \\
\midrule
canary-1b-v2   & 14.46 & & 6.82 & \\
Large-v3 Tuned & 11.6 & (\gooddelta{2.86}) & 5.5 & (\gooddelta{1.32}) \\
\midrule
parakeet-v3    & 8.83 & & 10.07 & \\
Large-v3 Tuned & 11.6 & (\baddelta{2.77}) & 5.5 & (\gooddelta{4.57}) \\
\bottomrule
\end{tabular}
\vspace{2mm}
\caption{WER comparison of our fine-tuned Whisper Large-v3 versus NVIDIA Canary and Parakeet models on Slovak benchmarks. Differences shown in parentheses relative to the baseline model (green = our model better, red = baseline better). Note: FLEURS and Common Voice v7 were included in NVIDIA models' training data.}
\label{tab:wer-large-canary-parakeet}
\end{table}

\section{Released Resources}
We present SloPal, a comprehensive Slovak parliamentary resource with three integrated components:

\textbf{SloPal Text Corpus:} 330,000 speaker-segmented transcripts (66M words, 220M tokens) spanning 2001--2024, providing coverage of all successfully retrieved parliamentary proceedings. Each segment includes speaker metadata (first name, last name, parliamentary role), session information (number, date, type), and transcriber notes. This structure enables applications beyond ASR: political discourse analysis, speaker-specific language modeling, diachronic linguistic studies, and parliamentary rhetoric research. The comprehensive metadata and speaker-level segmentation distinguish SloPal as the most comprehensive Slovak parliamentary text resource.

\textbf{SloPalSpeech Audio Dataset:} 2,806 hours of aligned audio-transcript pairs with 30-second segments optimized for Whisper training. While audio coverage (68.5\%) is lower than EuroSpeech's 89\%, the complete textual record remains available in SloPal.

\textbf{Fine-Tuned ASR Models:} Four Whisper variants (244M to 1.5B parameters) achieving 32--70\% WER reductions on Slovak benchmarks, with fine-tuned small model (244M parameters) approaching base large-v3 (1.5B parameters) performance at 6$\times$ fewer parameters, enabling practical deployment for Slovak ASR.

\paragraph{Licensing and availability.} The source parliamentary recordings and transcripts are public government documents with no copyright restrictions. We release the SloPal text corpus and SloPalSpeech aligned audio under the CC~BY~4.0 license, and the fine-tuned Whisper models under the same license as the original Whisper models (MIT). All resources are available on Hugging Face.\footnote{\url{https://huggingface.co/collections/NaiveNeuron/slopal}}

\section{Conclusion}
We presented SloPal, the most comprehensive Slovak parliamentary resource to date, combining 330,000 speaker-segmented transcripts with aligned speech data and fine-tuned ASR models. The dual-resource design addresses diverse research needs: the complete textual record supports NLP and discourse analysis, while 2,806 hours of aligned audio enable ASR development. Our speaker-segmented structure with rich metadata (names, roles, session info) enables parliamentary discourse analysis, political rhetoric studies, and diachronic linguistic research unavailable with audio-only resources. Fine-tuned Whisper models achieve up to 70\% WER reduction, with fine-tuned small model approaching base large-v3 performance at 6$\times$ fewer parameters.

We release all resources publicly: SloPal text corpus (330,000 segments, 66M words), SloPalSpeech aligned audio (2,806 hours), and four fine-tuned Whisper models\footnote{The text corpus includes all successfully parsed transcripts. SloPalSpeech contains the subset (2,806 hours) where audio was successfully aligned.}, providing comprehensive resources for Slovak NLP, ASR, and computational social science research.

\section{Future Work}
Parliamentary proceedings in dozens of national legislatures worldwide offer transcribed speech in low-resource languages. We plan to apply our anchor-based alignment pipeline to additional Slavic languages where parliamentary corpora exist but lack Whisper-optimized alignment. We also intend to investigate parameter-efficient fine-tuning (e.g., LoRA) to preserve multilingual capability while adapting to individual languages. We encourage other research groups to replicate our pipeline for their target languages using the publicly available code and data.

\subsection*{Limitations}
Whisper models are known to hallucinate due to their architecture. While the base model often repeated words like "Ďakujem," our fine-tuned models occasionally produced sequences such as "Pán poslanec," reflecting a bias toward parliamentary speech. We recommend adopting an approach similar to Faster Whisper \cite{systran_faster-whisper}, where compression ratio checks during inference can trigger a re-decode with adjusted temperature to mitigate hallucinations. Additionally, when encountering unknown words or phrases, the model may substitute them with parliamentary terms. For example, in one Common Voice sample with the transcript "ostrov Nihau" the model predicted "ostrov Mihál" where "Mihál" is a former politician's surname. Similar substitutions are expected in domains with vocabulary outside the parliamentary register.

Since fine-tuning updated all model parameters, the models specialize toward Slovak at the cost of other languages---a known trade-off of full-parameter fine-tuning. Parameter-efficient methods such as LoRA could preserve multilingual capability, which we leave to future work.

\section*{Acknowledgements}
We thank VÚB Banka for their generous support and for providing the GPU resources necessary to accomplish this work.
This research was partially funded by the EU NextGenerationEU through the Recovery and Resilience Plan for Slovakia under the project No.~09I02-03-V01-00029.

\section*{Bibliographical References}\label{sec:reference}

\bibliographystyle{lrec-coling2024-natbib}
\bibliography{lrec-coling2024-example}


\end{document}